\documentclass[10pt,twocolumn,letterpaper]{article}

\usepackage{iccv}
\usepackage{times}
\usepackage{epsfig}
\usepackage{graphicx}
\usepackage{amsmath}
\usepackage{amssymb}
\usepackage{units}

\usepackage{booktabs}
\usepackage{paralist}
\usepackage{multirow}
\usepackage{rotating}
\usepackage{afterpage}
\usepackage[table]{xcolor}
\usepackage{makecell}
\usepackage{algorithm}
\usepackage[noend]{algpseudocode}
\usepackage{array}
\usepackage{microtype}
\usepackage{multicol}

\usepackage{cite}
\usepackage[pagebackref=true,breaklinks=true,letterpaper=true,colorlinks,bookmarks=false]{hyperref}

\DeclareMathOperator*{\argmin}{arg\,min}
\usepackage[capitalize]{cleveref}
\crefname{section}{Sec.}{Secs.}
\Crefname{section}{Section}{Sections}
\Crefname{table}{Table}{Tables}
\crefname{table}{Tab.}{Tabs.}

\iccvfinalcopy % *** Uncomment this line for the final submission

\def\iccvPaperID{****} % *** Enter the ICCV Paper ID here

\ificcvfinal\fi

\begin{document}

\ificcvfinal\thispagestyle{empty}\fi

\newcommand{\density}{\sigma} 
\newcommand{\vx}{\boldsymbol{x}}  
\newcommand{\vd}{\boldsymbol{d}}
\newcommand{\vc}{\boldsymbol{c}}
\newcommand{\vr}{\boldsymbol{r}}
\newcommand{\vv}{\boldsymbol{v}}
\newcommand{\rthree}{\mathbb{R}^3}
\newcommand{\rfeature}{\mathbb{R}^m}
\newcommand{\rone}{\mathbb{R}}
\newcommand{\pcloud}{\mathcal P}
\newcommand{\vp}{\mathbf p}
\newcommand{\vf}{\mathbf f}
\newcommand{\nerfloss}{L_{\text{rgb}}}

%%%%%%%%% TITLE
\title{ParticleNeRF: A Particle-Based Encoding\\%
for Online Neural Radiance Fields}

\author{Jad Abou-Chakra $^1$
\and
Feras Dayoub $^2$
\and
Niko S\"{u}nderhauf $^1$
\and 
$^1$ Queensland University of Technology\\
$^2$ University of Adelaide
}

\maketitle

%%%%%%%%% ABSTRACT
\begin{abstract}
While existing Neural Radiance Fields (NeRFs) for dynamic scenes are offline methods with an emphasis on visual fidelity, our paper addresses the \textbf{online} use case that prioritises real-time adaptability. We present ParticleNeRF, a new approach that dynamically adapts to changes in the scene geometry by learning an up-to-date representation online, every \unit[200]{ms}. 
ParticleNeRF achieves this using a novel particle-based parametric encoding. We couple features to particles in space and backpropagate the photometric reconstruction loss into the particles' position gradients, which are then interpreted as velocity vectors. Governed by a lightweight physics system to handle collisions, this lets the features move freely with the changing scene geometry. We demonstrate ParticleNeRF on various dynamic scenes containing translating, rotating, articulated, and deformable objects.
ParticleNeRF is the first online dynamic NeRF and achieves fast adaptability with better visual fidelity than brute-force online InstantNGP and other baseline approaches on dynamic scenes with online constraints. 
Videos of our system can be found at our project website \href{https://sites.google.com/view/particlenerf}{https://sites.google.com/view/particlenerf}.
\end{abstract}

\vspace{-0.4cm}

\begin{figure}[t]
    \centering
    \includegraphics[width=0.85\linewidth]{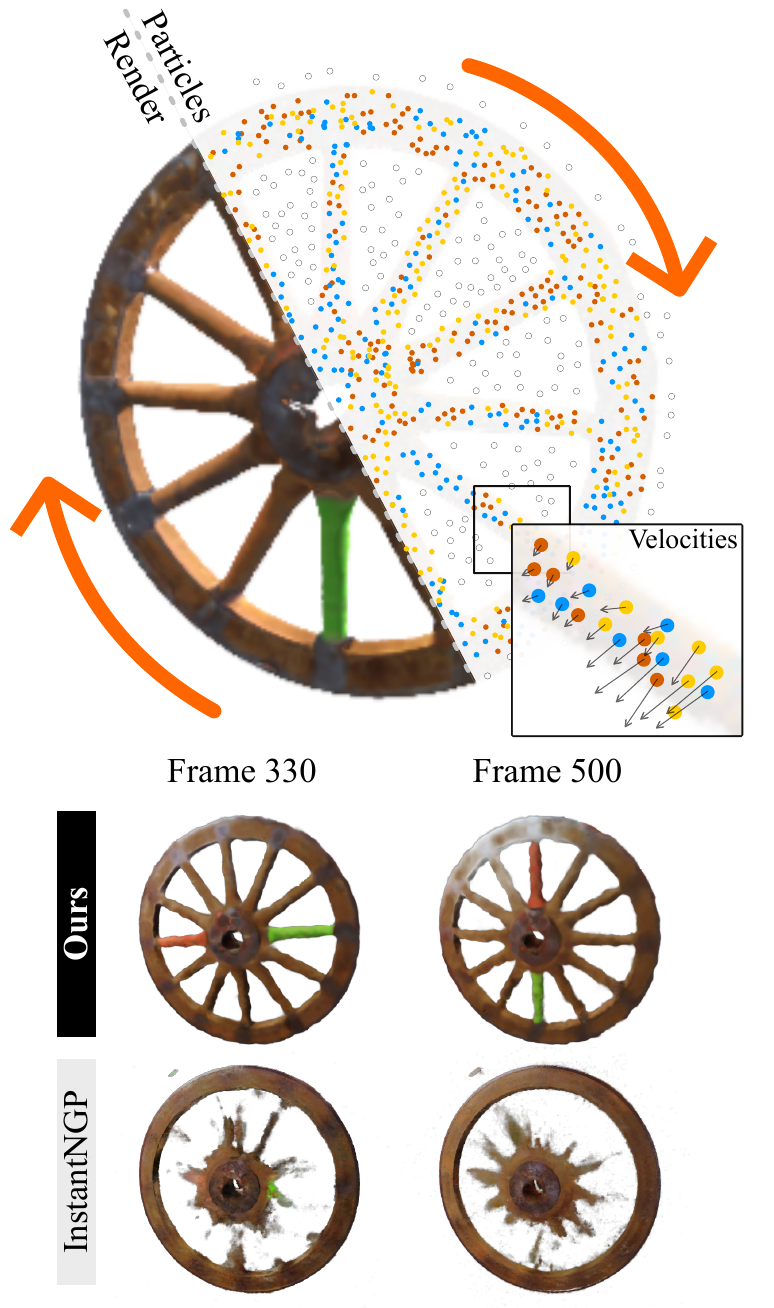}
    \caption{
    Our particle encoding significantly improves the ability to adapt to changing scenes when compared to online InstantNGP~\cite{mueller2022instant}, the top-performing parametric encoding. The particle encoding fills the space with particles at random locations, associating each particle with a feature. Both the positions of the particles and their features are optimized during training. As the wheel spins, the particles move to maintain a low reconstruction error. 
    }
    \label{fig:hero}
    \vspace{-0.6cm}
\end{figure}

\label{sec:intro}
\section{Introduction}

Neural Radiance Fields~\cite{mildenhall2020nerf} are 3D scene representations trained from images using a differentiable rendering process that allows them to render scenes from novel viewpoints with high visual fidelity.

Dynamic NeRFs~\cite{pumarola2020d, song2022nerfplayer, park2021hypernerf, park2021nerfies, gan2022v4d, tineuvox} learn to represent \emph{dynamic} scenes. 
However, these methods are currently \emph{offline}, which means
they require access to the \emph{entire} image sequence during training and can take several minutes or even hours to train for sequences that last only seconds in real-time.

Our paper addresses the challenge of \emph{\textbf{online}} dynamic Neural Radiance Fields that can continuously learn an up-to-date implicit 3D representation of the dynamic scene in real-time, without the need for access to future frames during training.

\begin{figure}
    \centering
    \includegraphics[width=0.85\linewidth]{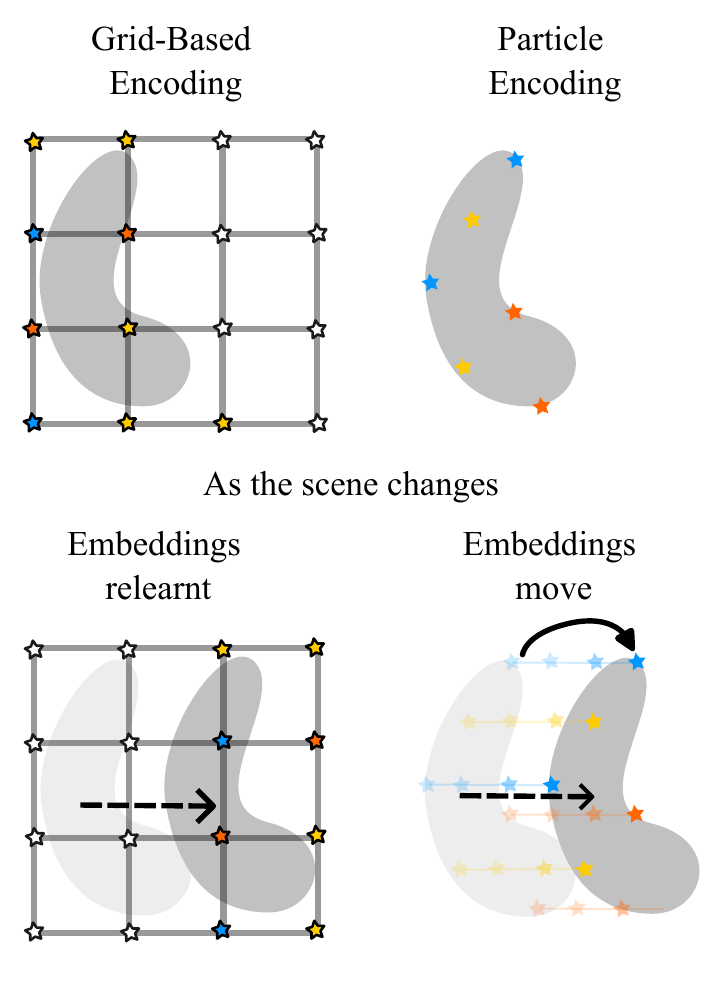}
    \caption{
    In a rigid grid-based encoding (e.g. InstantNGP~\cite{mueller2022instant}), features are associated with fixed positions in space and cannot move. If the scene geometry changes, features have to be unlearned and relearned in different positions. In contrast, our particle encoding associates features with particles that can move freely in 3D space. As the scene geometry changes, the particles can move to accommodate the scene change, resulting in faster adaptation and higher quality representation of dynamic scenes.
    }
    \label{fig:particle_vs_grid}
    \vspace{-0.4cm}
\end{figure}

We introduce ParticleNeRF, a novel dynamic NeRF approach that uses a particle-based encoding to represent the dynamic scene. Our key insight and novelty -- illustrated in Fig.~\ref{fig:hero} -- is that backpropagating the reconstruction loss into both the positions and the features of the particles  enables the embedding features to move in space as the scene changes. As we illustrate in Fig.~\ref{fig:particle_vs_grid}, this lets ParticleNeRF elegantly adapt to moving, articulated or deforming objects by adjusting the position of the feature-carrying particles. 

ParticleNeRF uses a simple but effective physics system to govern the particles' positions and velocities, interpreting the individual position gradients as velocity vectors. The physics system implements collision constraints that prevent the particles from accumulating too close to each other. This creates a soft upper limit on the number of nearest neighbours to be considered for interpolation when calculating the feature for a query point, thereby supporting ParticleNeRF's efficient training process.

In contrast to the time-conditioned deformation networks used in~\cite{pumarola2020d, song2022nerfplayer, park2021hypernerf, park2021nerfies, tineuvox, gan2022v4d}, which are designed for \emph{offline} training, ParticleNeRF incrementally adapts to changes in the environment every \unit[200]{ms} -- making it suitable for \emph{online} usage.

ParticleNeRF is the first method to address the challenge of \emph{online} learning of dynamic scenes (including rigid body transforms, articulated objects, and deforming objects). Our experiments show that ParticleNeRF can represent dynamic scenes with higher fidelity than (i) brute-force online InstantNGP that simply continuously trains the network as new frames become available, (ii) and other baselines (Section~\ref{sec:results}). Our ablation in Section~\ref{sec:ablation} shows that this superior performance on dynamic scenes is indeed due to the ability to backpropagate into the positions and move the feature-carrying particles during the training process. We furthermore show that ParticleNeRF's performance degrades gracefully with a decrease in the number of particles and the nearest neighbour search radius. 

In summary, our contributions are:
\begin{compactenum}
    \item A novel online NeRF formulation that can adapt to the changes in dynamic scenes every \unit[200]{ms} without using time-conditioned deformation networks;
    \item A new particle-based encoding that associates features with moving particles and allows the gradients from the photometric loss to propagate to and change the positions of the particles;
    \item The incorporation of a physics system into the NeRF formulation to update the motion of the particles while preventing collisions.
\end{compactenum}

Our implementation will be made available upon publication. Videos of ParticleNeRF rendering dynamic scenes can be found on our project website \href{https://sites.google.com/view/particlenerf}{https://sites.google.com/view/particlenerf}.

\section{Related Work}
\label{sec:related}
We review NeRFs in the context of (i) how they have been used to tackle dynamic environments, (ii) how certain encodings allow close to realtime training and rendering, and lastly (iii) how our particle encoding is different to three other point-based encodings suggested to date.

\vspace{-0.5cm}
\paragraph{Dynamic NeRFs}

Several papers extended the NeRF model to encompass dynamic scenes~\cite{pumarola2020d, song2022nerfplayer, park2021hypernerf, park2021nerfies, tineuvox, gan2022v4d, Xian_2021_CVPR, Gao_2021_ICCV, Li_2022_CVPR}. The distinguishing factor between the static and dynamic NeRF formulations is that the first represents a single static scene while the latter encodes the evolution of a 3D scene over time. During training, these dynamic NeRFs can access images taken at any time step. They are therefore designed to be trained \textbf{after} a sequence of images has already been taken. To date, the majority of these works are not close to being realtime capable taking minutes and even hours to train. Since these dynamic NeRF approaches use the complete set of images during training, we refer to them as ``offline'' methods. Alternatively, we seek to create a NeRF at every timestep and only capture the latest state of the scene. In this context, our method is ``online''. For many applications, representing a time-varying scene within the NeRF itself is unnecessary. A robot, for example, is usually only interested in the latest state of the world. 

Learning an online NeRF is therefore the challenge of learning a NeRF at a timestep $t$ given a NeRF at a timestep $t-1$ and a set of images taken at time $t$. Our particle encoding is formulated to tackle this problem.

\vspace{-0.4cm}
\paragraph{Embeddings}
In general, a NeRF is composed of two main parts: the encoding, and a multi-layered perceptron (MLP). An encoding transforms the spatial input of the NeRF to an alternative feature space where learning can be made easier~\cite{tancik2020fourfeat}. The MLP transforms the resultant feature into values that represent color and geometry. Mildenhall et al.~\cite{mildenhall2020nerf} introduced the first version of NeRFs with a Fourier encoding which is a simple trigonometric map. The Fourier encoding does not have any parameters that are learnt as part of the training and is not associated with a discrete datastructure. It is, therefore, referred to as non-parametric, or functional. This version of NeRF takes hours to train because it requires a large MLP to represent the scene. Later works~\cite{mueller2022instant,sun2022direct,Chen2022ECCV} show that by storing a large set of features in a discrete datastructure, the spatial input could be used to index into and interpolate between features in that set. The computed features become the embeddings that are consumed by the NeRF's neural network. Since the features in the set are stored as parameters and are updated during training, these are referred to as ``parametric'' encodings by~\cite{mueller2022instant}. By storing features in memory, the computational burden on the MLP is reduced and training speed is significantly increased. The strategy used to index and interpolate into the feature set distinguishes the various types of parametric encodings.  Almost all the parametric encodings conceptually rely on a fixed grid datastructure~\cite{mueller2022instant, sun2022direct, Chen2022ECCV} with the exception of~\cite{xu2022point} which uses a point-based encoding. The grid encompasses the workspace of the NeRF and each node within is associated with a feature. Therefore, while the features associated with the nodes can be learnt, they cannot be moved. 

In this paper, we show that if the features are allowed to be both learnt and moved, then NeRFs can be made to incrementally adapt to dynamic scenes. We introduce a particle encoding, where features are associated with moving points in space. We show that in a changing scene, the feature-holding particles move in accordance with the scene -- \Cref{fig:particle_vs_grid}. 

\vspace{-0.5cm}
\paragraph{Point-based NeRFs}
There are three works that use points as part of their formulation. Point-NeRF and SPIDR~\cite{xu2022point, liang2022spidr} utilize points as a data structure to store features, but their embedding differs significantly from ours. Notably, they do not address the dynamic scene use case and they they do not allow their points to move during optimization. In addition, they require depth data, their training takes in the order of minutes to hours, and their encoding is not invariant to $\text{SE}(3)$ transformations. 
NeuroFluid~\cite{guan2022neurofluid} uses particles and NeRFs to specifically model the behaviour of fluids. Their encoding is non-parametric and their points represent particles of a physical medium rather than descriptors of an embedding space.

\begin{figure*}[t]
    \centering
    \includegraphics[width=0.9\linewidth]{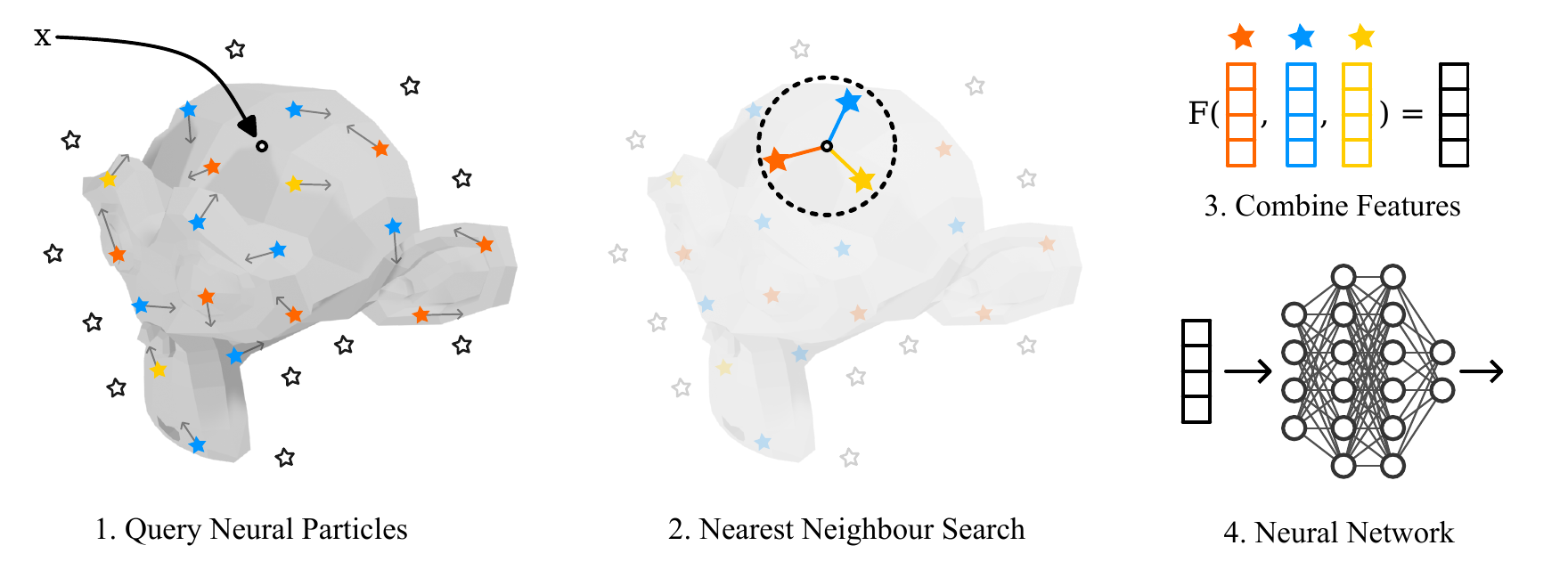}
    \caption{
    Illustration of our particle encoding. \textbf{(1)} A query point $\vx$ is sampled in space. \textbf{(2)} The features and positions of the particles within a search radius are retrieved. \textbf{(3)} The features and distances from the query point are used to interpolate the feature at the query point $\vx$. \textbf{(4)} The resulting feature is evaluated by the neural network to give color and density. To train the encoding, the loss gradients are backpropagated through the network \textbf{(4)}, the query feature \textbf{(3)}, and finally into the positions and features of the particles in \textbf{(2)}.
    }
    \label{fig:method}
\end{figure*}

\section{NeRF Preliminaries}
\label{sec:review}
A NeRF is a continuous representation of a 3D scene that maps a point $\vx \in \rthree$ and a unit-norm view direction $\vd \in \rthree$ to a color $\vc\in\rthree$ and a density value $\sigma \in \rone$. During training, rays are cast from camera centers through their image plane. Each ray $\vr$ is associated with a direction vector $\vd$ and a set of points $\vx_i$ progressively sampled from the ray's length~\cite{mildenhall2020nerf}. The points $\vx_i$ and the direction vector $\vd$ are mapped by the NeRF to $\vc_i$ and $\density_i$. Defining $\delta_i = \lVert \vx_{i+1} - \vx_i \rVert$, the expected color of the ray $\hat{C}(\boldsymbol{r})$ is given by:

\vspace{-0.2cm}

\begin{equation}
    \hat{C}(\boldsymbol{r}) = \sum_{i=1}^{N}w_i\boldsymbol{c}_i 
\end{equation}

\vspace{-0.5cm}
\begin{equation*}
w_i = T_i(1-\exp(-\sigma_i\delta_i)) \quad \text{and} \quad T_i = \exp{\left(-\sum_{j=1}^{i-1}{\sigma_j\delta_j}\right)}
\end{equation*}

A photometric loss is defined as
\begin{equation}
    L_{\text{rgb}} = \sum_{\boldsymbol{r} \in \mathcal{R}}\| \hat{C}(\vr) - \vc_{\text{gt}}(\vr) \|_{2} 
    \label{eq:loss}
\end{equation}

The ground-truth color $\vc_{gt}(\vr)$ is the color of the pixel in the training image that the ray $\vr$ intersects. $\mathcal{R}$ is the set of all rays that pass through the training images. Refer to~\cite{tagliasacchi2022volume, 468400} for an in-depth derivation.

\vspace{-0.4cm}
\paragraph{Encodings} 
Given a point $\vx_i$ and a direction vector $\vd_i$, a NeRF first maps each to an embedding space $E_x(\vx_i)$ and $E_d(\vd_i)$. The embeddings are subsequently consumed by a multilayered perceptron to output color and density. Not including an embedding layer critically deteriorates ability of the NeRF to reconstruct scenes~\cite{tancik2020fourfeat}. The choice of the embedding layer can significantly reduce the size of the MLP required~\cite{Chen2022ECCV, mueller2022instant, xu2022point} and in some cases removes the need for one entirely~\cite{Fridovich-Keil_2022_CVPR}. Close to realtime performance is possible due to a particular class of encodings that the authors of~\cite{mueller2022instant} refer to as ``parametric''. These encodings are characterized by a discrete datastructure that can be queried by the incoming coordinate $\vx_i$ to produce a set of features $\vf_j$. The features $\vf_j$ are then combined to produce an embedding $\vf_i$. By storing the features in memory, a large MLP is no longer required and significant savings are made in both rendering and training times when compared to other formulations that use functional encodings~\cite{mildenhall2020nerf, verbin2022refnerf, barron2021mip, barron2022mip, chng2022garf}.

\section{ParticleNeRF}
ParticleNeRF introduces a new particle-based encoding that associates features with moving particles and allows the gradients from the photometric loss to propagate through and change the positions of the particles. This is a novel method of dealing with dynamic scenes in an online manner without using time-conditioned deformation networks.  

Our particle encoding is formulated as a neural pointcloud $\pcloud = \{(\vx_i, \vv_i, \vf_i) : i=1,2,..., M \}$, where $\vx_i \in \rthree$ is the position of a latent particle in the system, $\vv_i \in \rthree$ is the velocity of the particle, $\vf_i \in \rfeature$ is its associated feature, and $M$ is the total number of particles. The pointcloud $\pcloud$ is an irregular grid of latent features that defines the map F from a coordinate $\vx_j \in \rthree$ to $\vf_j$.

 \begin{equation}
  \vf_j = F(\vx_j, \pcloud) = \sum_{(\vx_i, \vf_i) \in \pcloud}{w(\lVert \vx_j - \vx_i \rVert_2 )\vf_i}
  \label{eq:feature}
\end{equation}

where $w$ is the bump function -- a compactly supported radial basis function (RBF) -- given by:

 \begin{equation}
w(r) = \begin{cases}
\exp\left({-\displaystyle \frac{s^2}{s^2-r^2}}\right),  &\text{for }r\in (-s^2, s^2)\\
0, & \text{otherwise}
\end{cases}
\end{equation}

The search radius $s$ controls the influence of a particle on a region of space. The compact RBF decreases the influence of a given particle on the interpolated feature to $0$ as its distance approaches the search radius. This allows for the efficient calculation of $\vf_j$ with a fixed-radius nearest neighbour search algorithm~\cite{fixrnn} where the radius is set to $s$. We find that using unnormalized features encourages the particles to spread across the geometry.

ParticleNeRF uses the same architecture as InstantNGP (a 3 layer MLP with 64 neurons each), but replaces its hash encoding with our particle encoding. Note that if a query point finds no neighbouring particles within the search radius $s$, the feature is set to $\boldsymbol{0}$. \Cref{fig:method} illustrates our method. 

We optimize the MLP parameters $\Phi$, the particle features $\vf_i$ and the particle positions $\vx_i$ relative to the NeRF loss $\nerfloss$ defined in \cref{eq:loss}:

\begin{equation}
    \Phi^*, \{\vf_i\}^*, \{\vx_i\}^* = \argmin_{\Phi, \{\vf_i\}, \{\vx_i\}} L_{\text{rgb}}
\end{equation}

A core novelty of our approach is that the gradients of the NeRF loss are used to update \emph{both} the position and the associated features of the particles. This allows the particles to move with the geometry in the scene as it changes, leading to a higher fidelity representation of dynamic scenes. As our experiments in Section~\ref{sec:results} will show, ParticleNeRF can adapt to changes in the scene \emph{online}, i.e. every \unit[200]{ms}, which is orders of magnitude faster than previous NeRFs in dynamic scenes~\cite{park2021hypernerf, park2021nerfies, pumarola2020d, Xian_2021_CVPR, Gao_2021_ICCV, Li_2022_CVPR}, and enables a much higher reconstruction quality compared to online InstantNGP~\cite{mueller2022instant} and other baselines.

\subsection{Position Based Dynamics}
We observe that propagating the NeRF loss into the particle positions can sometimes lead to multiple particles accumulating in a small region of space. Although this does not adversely affect the reconstruction quality, it is not desirable because it needlessly increases the number of neighbours that each query point has to process during training and evaluation. To address this problem in a structured way, we build a position-based physics system~\cite{pbd, xpbd} into InstantNGP that resolves the dynamics of our particles. 

Position-based dynamics (PBD) is a simple, robust, and fast physics system that can evolve the motion of our particles while resolving the constraints we put on them. We interpret the gradients $\frac{d\nerfloss}{d\vx_i}$ as scaled velocity vectors that are added to the particles' current velocity -- \Cref{alg}. Using PBD not only provides a means of limiting the number of neighbours a particle has, but it also creates future opportunities to add other constraints into the system. For example, if an object is known to be a rigid object, or part of an articulated body, the particles can be suitably constrained.

\begin{algorithm}[tb]
\caption{PBD Physics Step}
\label{alg}
\begin{algorithmic}[1]
\For{all particles $i$} 

\State $\vv_i \gets \gamma \vv_i - \alpha\displaystyle\frac{d\nerfloss}{d\vx_i}$ \Comment{Update velocities}
\State $\vp_i \gets \vx_i$  \Comment{Store previous positions} 
\State $\vx_i \gets \vx_i + \Delta t \vv_i$ \Comment{Integrate positions} 

\EndFor

\For{all collision pairs i, j} 
    \State $l \gets ||\vx_j - \vx_i||$
    \State $\vx_i \gets \vx_i + 0.5(l-\delta)\displaystyle\frac{\vx_j - \vx_i}{l}$ \Comment{Resolve collisions} 
\EndFor

\For{all particles $i$} 
\State $\vv_i \gets (\vx_i - \vp_i) / \Delta t$ \Comment{Update velocities} 
\EndFor
\end{algorithmic}
\end{algorithm}

In all our experiments, we choose a damping factor $\gamma = 0.96$, a timestep $\Delta t = 0.01$, a minimum distance of $\delta = 0.01$. The base InstantNGP implementation transforms a given scene so that it fits within a unit cube. Our minimum distance $\delta$ in the physics system and the search radius $s$ that is used to query neighbouring particles are given in the units of the unit cube.
 
\subsection{Implementation}
We build our particle encoding as an extension of the InstantNGP implementation released by M\"uller \etal ~\cite{mueller2022instant}. We clip the particle position gradients so that their norms do not exceed the search radius. We implement the sorting-based fixed-radius nearest search algorithm described by~\cite{green2010particle} and the position-based dynamics physics system described by~\cite{pbd} as additional CUDA kernels~\cite{cuda} within InstantNGP and will make the implementation available after publication. Two different Adam optimizers~\cite{kingma2014adam} are used. The first is for the parameters of the MLP $\Phi$ and the second is for the particle features $\vf_i$. Both are parameterized by $\beta_1 = 0.9$, $\beta_2 = 0.99$, $\epsilon = 10^{-10}$ with the learning rate set at $0.01$. The particle positions are updated through the PBD system where collision constraints can be resolved.

\iffalse
\begin{figure*}[t]
    \centering
    \includegraphics[width=\linewidth]{Figures/SyntheticVisuals.pdf}
    \caption{
    We test our encoding on an animated version of the Blender dataset~\cite{mildenhall2020nerf}. \textbf{(a)} shows an object rotating around its up axis. \textbf{(b)} shows an object translating from side to side. InstantNGP cannot learn features fast enough to maintain a high quality reconstruction. ParticleNeRF is able to move its features in space and maintain the structure of the object.
    }
    \vspace{0.2cm}
    \label{fig:synth_vis}
\end{figure*}
\input{synthetic_table.tex}
\fi

\section{Experiments}
\label{sec:results}
We experimentally evaluate ParticleNeRF and show its abilities to represent a dynamic environment online.

\noindent\textbf{Datasets:} 
We construct a dynamic dataset consisting of 6 scenes, inspired by D-NeRF~\cite{pumarola2020d}, comprising of scenes with deformable and articulated moving objects that loop. In addition, we created an animated version of the original NeRF Blender dataset~\cite{mildenhall2020nerf}, by applying translations and rotations to the objects.

We note that datasets such as~\cite{park2021hypernerf, park2021nerfies} that are commonly used to evaluate offline dynamic NeRFs are not compatible with ParticleNeRF's (and InstantNGP's) requirement for multiple cameras per frame. We discuss this limitation further in \cref{sec:LimAndConc}.

\noindent\textbf{Baseline:}
We compare ParticleNeRF against online InstantNGP~\cite{mueller2022instant} and online TiNeuVox~\cite{tineuvox}. 

InstantNGP is the fastest NeRF implementation currently available, and it can be utilized for online applications by continuously learning on new frames without resetting the weights. We are not the first to propose this, as EvoNeRF~\cite{kerr2022evonerf} uses a continuous data stream to train InstantNGP for a robotic grasping task. InstantNGP's fast training speed allows us to evaluate whether a brute force approach of continuous learning is a feasible strategy for online dynamic scenes. 

As most of the dynamic NeRF approaches~\cite{park2021hypernerf, park2021nerfies, pumarola2020d, Xian_2021_CVPR, Gao_2021_ICCV, Li_2022_CVPR} published to date are offline methods that are not trained at the required speeds for online applications, we do not compare our approach against them. However, a recent offline method called TiNeuVox~\cite{tineuvox} has been released, which trains within minutes by utilizing a time-conditioned deformation network and voxel-based parametric encodings. We compare our approach against the two configurations of TiNeuVox released by the authors: TiNeuVox and TiNeuVox-S. We convert them to an online use-case where only images from the current timestep can be sampled. 

\noindent\textbf{Hyperparameters:}
For all the experiments involving the particle encoding, we initialize the unit cube containing the scene with a grid of uniformly spaced particles (We obtain comparable outcomes when initializing the particles randomly). Each particle is associated with a 4 dimensional feature and is initialized randomly between $\pm10^{-2}$. Increasing the dimension of the feature has minimal impact. 

\begin{figure}[t]
    \centering
    \includegraphics[width=0.95\linewidth]{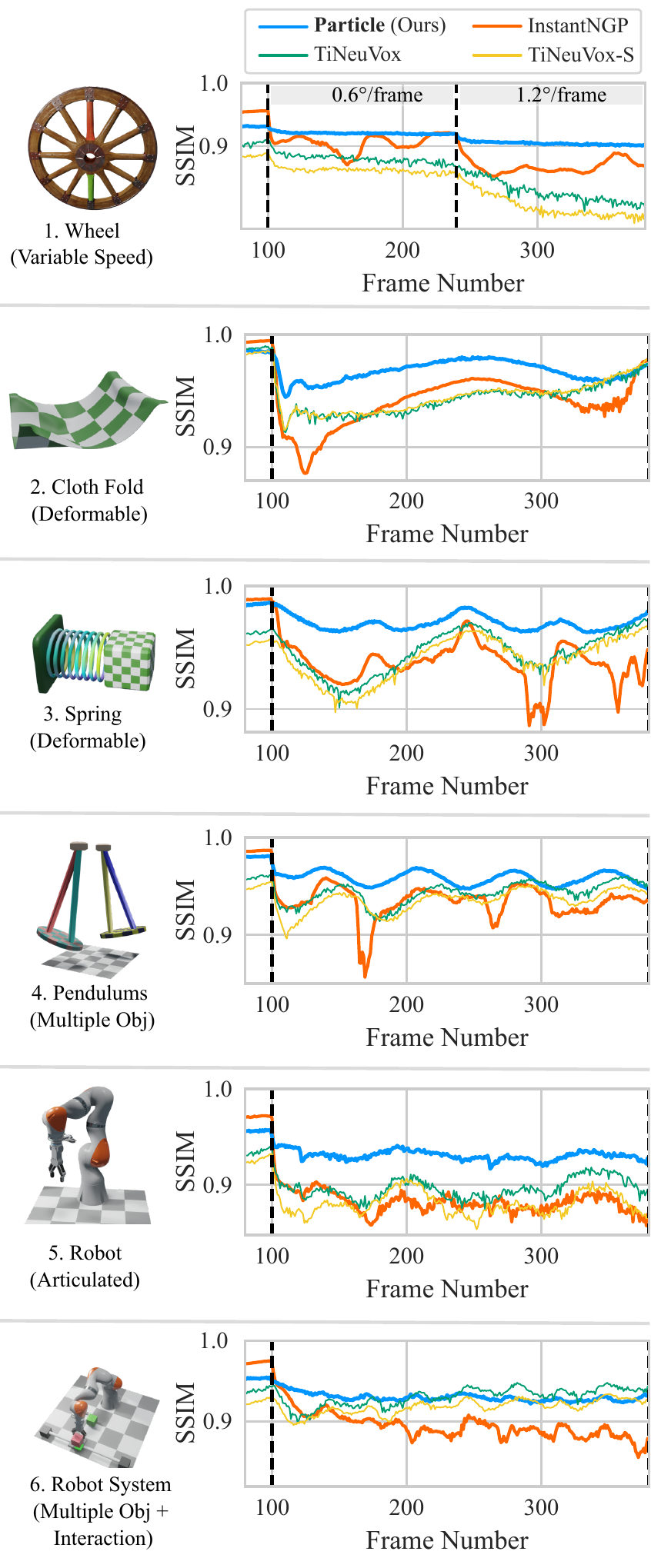}
    \caption {
    ParticleNeRF evaluated on 6 scenes with deformable, articulated, and multiple object movements. Scenes are held static for 100 frames before the animation begins. ParticleNeRF provides a stable representation during movements. InstantNGP can recover when motions slow down. Online TiNeuVox fails when movements get faster. 
    }
    \label{fig:dyn_exps}
\end{figure}

\begin{figure}[t]
    \centering
    \includegraphics[width=\linewidth]{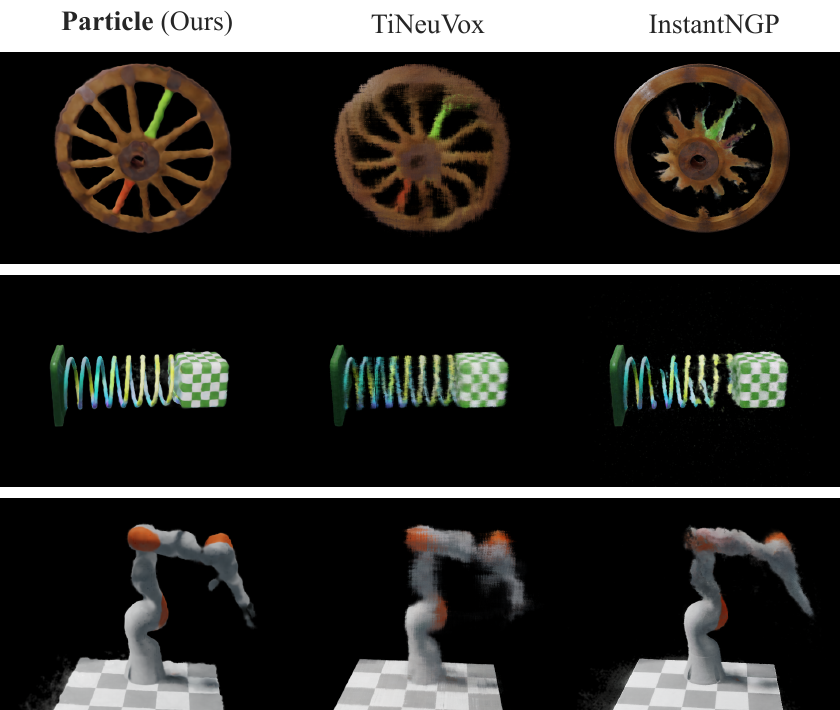}
    \caption {
    Image captures from a novel viewpoint at the 300th frame for ParticleNeRF, TiNeuVox, and InstantNGP. 
    }
    \label{fig:dyn_visuals}
   
\end{figure}

\subsection{Dynamic Dataset}
\label{sec:moving_blender}
We evaluate our method using our dynamic dataset (\Cref{fig:dyn_exps}). Each scene in the dataset has one or multiple objects being looked upon by 20 cameras distributed on the top half of the surface of a sphere. The objects are either articulated (robot), deformable (spring, cloth), or rigid (wheel). The remaining scenes are comprised of multiple moving objects (pendulums, robot system). 

During each experiment, the scene remains fixed for 500 training steps, after which the animation starts. ParticleNeRF, online InstantNGP, and online TiNeuVox have only \textbf{five} training iterations before loading the next animation step. 

At the end of every frame, we evaluate the photometric reconstruction quality (SSIM, PSNR and LPIPS) from 10 unseen views. The SSIM values are plotted against each frame in \Cref{fig:dyn_exps}. The average PSNR, SSIM, and PSNR values of all the scenes for each method are shown in \Cref{tab:dyn_tab}. Qualitative visualizations are shown in \Cref{fig:dyn_visuals} and the full videos can be found on the \href{https://sites.google.com/view/particlenerf}{project page}. 

Our findings reveal that ParticleNeRF effectively adapts to motion within a small number of training steps, while the performance of online InstantNGP noticeably declines under motion. InstantNGP is able to recover when the speed of movements slow down. Nonetheless, ParticleNeRF exhibits reduced performance when objects in the scene have varying sizes, which is due to its reliance on a fixed search radius, as shown by the decreased performance in the robot system scene. Additionally, online TiNeuVox exhibits artifacts, particularly visible in the wheel scene, caused by the deformation network's inability to adapt within 5 training iterations.

\begin{table}[t]
\centering
\caption{Average quantitative performance of methods on all dynamic scenes\\}
\label{tab:dyn_tab}
\begin{tabular}{llll}
\toprule
 Method & PSNR$\uparrow$ & SSIM$\uparrow$ & LPIPS$\downarrow$ \\
\midrule
Particle & \textbf{27.47} & \textbf{0.94} & \textbf{0.08}\\
InstantNGP & 24.69 & 0.91 & 0.12 \\
TiNeuVox & 27.28 & 0.91 & 0.13 \\
TiNeuVox-S & 26.64 & 0.92 & 0.14 \\
\bottomrule
\end{tabular}
\vspace{-0.3cm}
\end{table}

\subsection{Animated Blender Dataset}
We further evaluate our method using a novel animated
version of the standard “Blender” dataset~\cite{mildenhall2020nerf}. Each of the 8 scenes is made to rotate or translate at a certain speed for 100 frames to test ParticleNeRF's ability to deal with elementary motions. At the end of every frame, we evaluate the photometric reconstruction quality (PSNR) from 7 unseen views. In total, 56 animations are tested and their results are summarized in \Cref{tab:anim_results}. Videos and a more detailed view of the results can also be found in the attached supplementary. The results reinforce our findings that allowing particles to move provides a method of adapting to moving geometries.  

\begin{table}[t]
\centering
\caption{Average PSNR on the animated blender scenes from~\cite{mildenhall2020nerf}.\\}
\resizebox{0.9\linewidth}{!}{%
\label{tab:anim_results}
\begin{tabular}{@{}p{0.75cm}>{\centering\arraybackslash}p{0.6cm}*{4}{>{\centering\arraybackslash}p{0.3cm}}*{3}{>{\centering\arraybackslash}p{0.3cm}}@{}}

\toprule
\multirow{2}{*}{Model} & \multirow{2}{*}{Enc.} & \multicolumn{4}{l}{Rot. ($^\circ$/frm)} & \multicolumn{3}{l}{Trans. (cm/frm)} \\
& & 1 & 2 & 3 & 4 & 1 & 2 & 3 \\
\midrule
\rowcolors{2}{gray!25}{white}
\multirow{2}{*}{Chair} & ngp & 21 & 18 & 17 & 17  & 23&	19&	18 \\
& ours & \textbf{24} & \textbf{23}  & \textbf{22}  & \textbf{21}  & \textbf{24}&	\textbf{23}&	\textbf{22}  \\
\multirow{2}{*}{Drums} & ngp & 19 &	18	&17	&16   & 19&	18&	17 \\
& ours & \textbf{20} &	\textbf{19}&	\textbf{19}	&\textbf{18}                                  &\textbf{20}&	\textbf{19}&	\textbf{19}   \\
\multirow{2}{*}{Ficus} & ngp &  22&	22&	21&	21      & 22&	21&	\textbf{21} \\
& ours & \textbf{23}&	\textbf{22}&	\textbf{22}&	\textbf{21}                        & \textbf{22}&	\textbf{22}&	21  \\
\multirow{2}{*}{Hotdog} & ngp &  26&	24&	23&	23     & 25&	23&	21 \\
& ours & \textbf{27}&	\textbf{27}&	\textbf{26}&	\textbf{25}  & \textbf{26}&	\textbf{25}&	\textbf{24}   \\
\multirow{2}{*}{Lego} & ngp &  21&	19&	18&	18              & 21&	20&	19 \\
& ours & \textbf{23}&	\textbf{22}&	\textbf{21}&	\textbf{21} & \textbf{23}&	\textbf{22}&	\textbf{21}   \\
\multirow{2}{*}{Mic} & ngp &  26&	24&	23&	23              & 27&	24&	23 \\
& ours & \textbf{28}&	\textbf{27}&	\textbf{26}&	\textbf{25}  &\textbf{28}&	\textbf{27}&	\textbf{26}  \\
\multirow{2}{*}{Ship} & ngp &  23&	22&	22&	21              & 22&	21&	21 \\
& ours & \textbf{23}&	\textbf{23}&	\textbf{23}& \textbf{22} & \textbf{23}&	\textbf{23}&	\textbf{22}   \\
\multirow{2}{*}{Matrl.} & ngp &  21&	19&	17&	17& 23 & 21 & 19 \\
& ours & \textbf{23} &	\textbf{23} &	\textbf{22} & 	\textbf{21}  & \textbf{23} & \textbf{23} & \textbf{23}   \\
\bottomrule
\end{tabular}
}
\begin{tabular}{cc}
\includegraphics[width=0.35\linewidth]{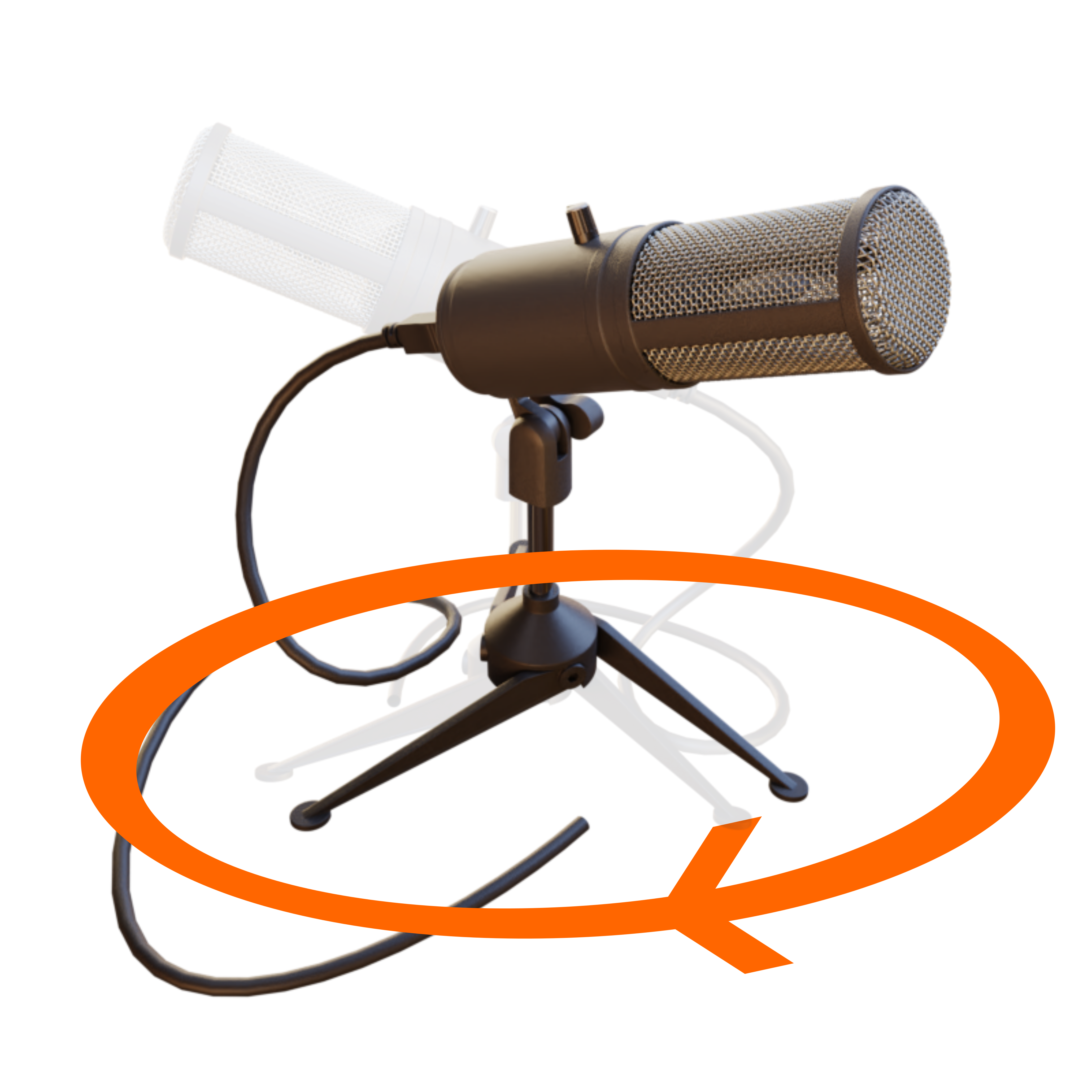} & \includegraphics[width=0.35\linewidth]{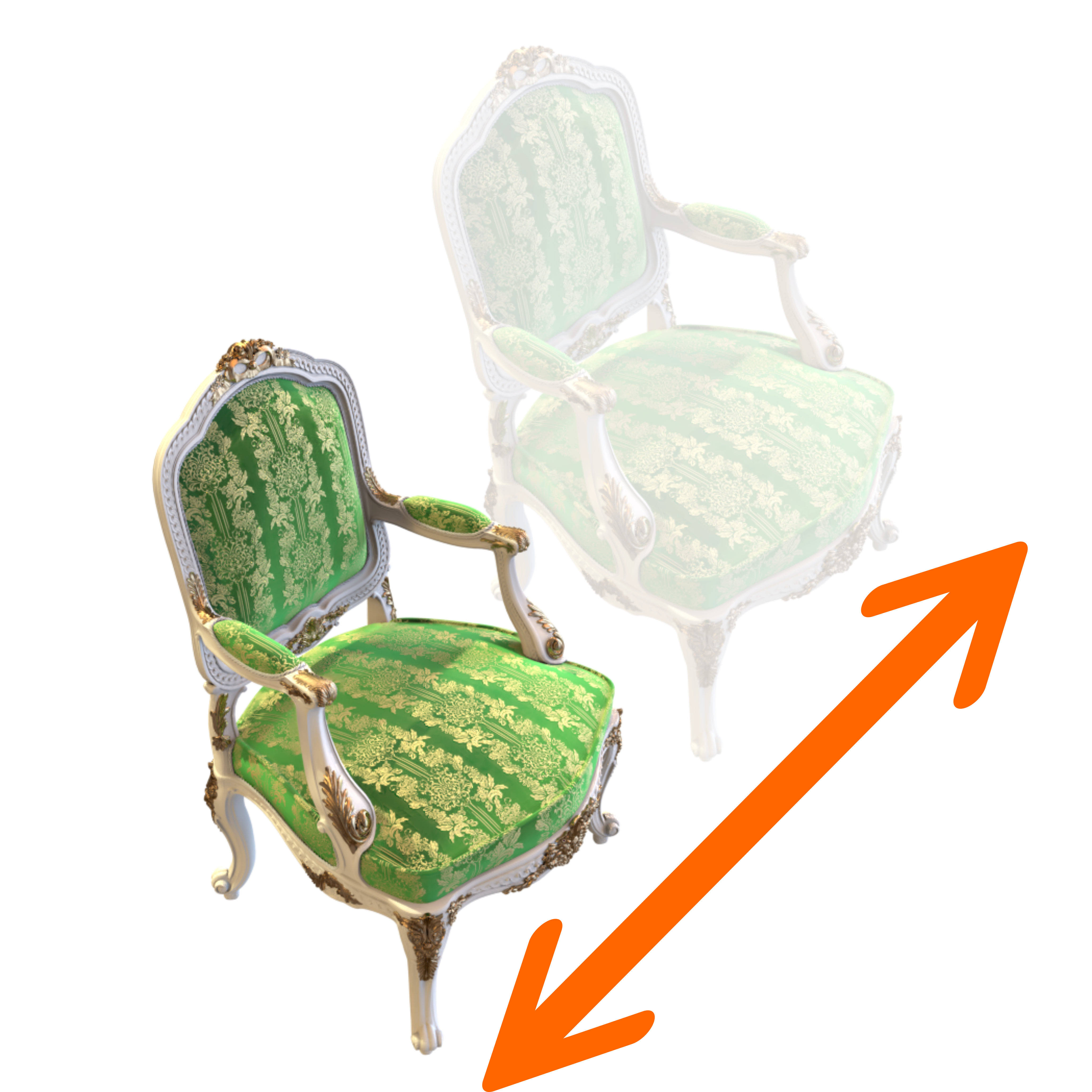} \\
(a) Rotation & (b) Translation \\
\end{tabular}
\vspace{-0.4cm}
\end{table}

\iffalse
\begin{table}[h]
\centering
\caption{Average quantitative performance of methods on all the animated blender scenes from~\cite{mildenhall2020nerf}.\\}
\label{tab:anim_results}
\begin{tabular}{lccccccc}
\toprule
\multirow{2}{*}{Model} & \multicolumn{4}{l}{Rot ($^\circ$/frame)} & \multicolumn{3}{l}{Trans (cm/frame)} \\

                       &                         1 & 2 & 3 & 4 & 1 & 2 & 3 \\
\midrule
                       
NGP             &                                &       &       &      &           &          &          \\
\textbf{Particle}                                    &       &       &       &      &           &          &         \\
\bottomrule
\end{tabular}
\begin{tabular}{cc}
\includegraphics[width=0.35\linewidth]{Figures/MicRot.png} & \includegraphics[width=0.35\linewidth]{Figures/ChairTrans.png} \\
(a) Rotation & (b) Translation \\
\end{tabular}
\end{table}
\fi

\begin{figure*}[t]
    \centering
    \includegraphics[width=\linewidth]{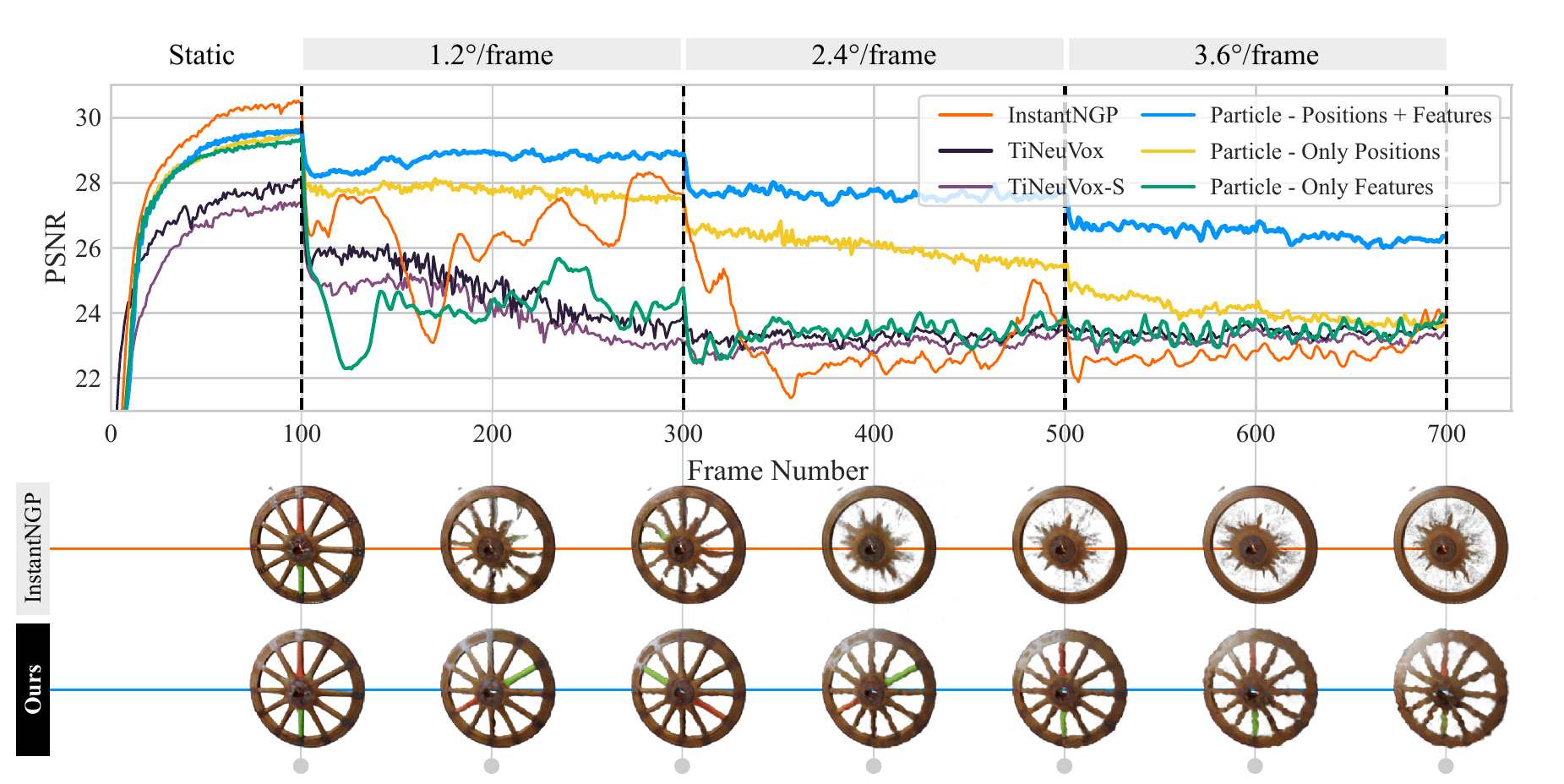}
    \caption{
On the Wheel dataset, ParticleNeRF maintains the structure of the wheel as it rotates. We show how our particle encoding performs in three settings: \textbf{(i)} When both the features and the positions of the particles are optimized (blue). \textbf{(ii)} When only the positions of the particles are optimized (yellow). \textbf{(iii)} When the only the particle features are optimized (green). Lastly we show how InstantNGP~\cite{mueller2022instant} performs (orange). We illustrate the results of the reconstruction from an unseen viewpoint at the frames indicated for both InstantNGP and the complete ParticleNeRF, corresponding to setup \textbf{(i)}. 
Even at the lowest PSNR at frame 700, our method preserves the wheel's structure.
    }
    \label{fig:wheel_part_vs_instantngp}
\end{figure*}

\label{sec:ablation}
\begin{figure*}[t]
    \centering
    \includegraphics[width=\linewidth]{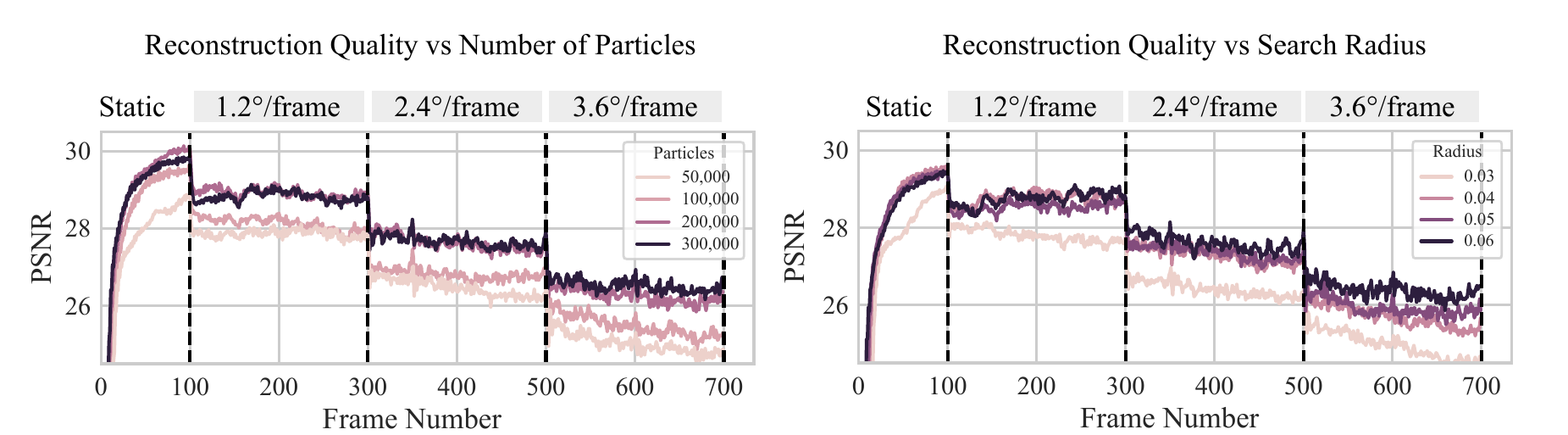}
    \caption{
An ablation showing the effect of the number of particles (left) and the particle search radius (right) on the photometric reconstruction quality. When ablating the number of particles, search radius is set at 0.04. When ablating the search radius, number of particles is set at 100,000. In both cases, there is a diminishing effect at higher speeds.
    }
    \label{fig:ablation}
\end{figure*}

\subsection{Ablation}
\label{sec:wheel_exp}
To thoroughly ablate and understand the dynamic representational ability of ParticleNeRF, we use the wheel scene from the dynamic dataset. To investigate ParticleNeRF's and the baselines' performance in static scenes and dynamic scenes with varying degrees of motion, we change the rotation speed of the wheel from static to successively faster speeds. Most of the following results are illustrated in Fig.~\ref{fig:wheel_part_vs_instantngp}.

Models are given 5 training steps per frame and are configured with the same parameters from \Cref{sec:moving_blender}.  

\vspace{-0.3cm}
\paragraph{Optimising Particle Positions}
We first investigate the contribution of ParticleNeRF's ability to backpropagate the rendering loss into the particle positions to the overall performance. In Fig.~\ref{fig:wheel_part_vs_instantngp} we show PSNR performance when allowing the loss to propagate into only the features (green); only the particle positions, keeping the features at their random initialisations (yellow), and allowing to learn both the positions and the features (blue). We make the following observations: First, when allowing to lean only the features (green), the performance is identical to InstantNGP (orange), especially at the higher motion rates. This is not surprising, since in that setup ParticleNeRF operates like InstantNGP with randomly positioned feature embeddings, instead of a fixed grid structure. Second, keeping the features fixed, but allowing the particles to move, leads to increased PSNR performance that already outperfoems the baselines in all but a few dynamic frames. Third, the combination of learning positions and features achieves best performance.

\vspace{-0.4cm}
\paragraph{Increasing Motion Magnitude}
From Fig.~\ref{fig:wheel_part_vs_instantngp} we can see that brute-force InstantNGP achieves a higher PSNR on static scenes, but drops significantly in dynamic scenes as it is not capable of relearning features fast enough to adapt.
While ParticleNeRF has a slightly worse performance on static scenes, it maintains its PSNR with only slight degradation as the scene becomes more dynamic. Comparing the renderings at the bottom of Fig.~\ref{fig:wheel_part_vs_instantngp}, we can see the significant motion artifacts exhibited by brute-force InstantNGP. We furthermore observe a ghosting effect where InstantNGP quickly recovers structure upon the wheel's return to its original configuration. This is manifested as oscillations in the PSNR metric in \Cref{fig:wheel_part_vs_instantngp}.

\vspace{-0.4cm}
\paragraph{Influence of Search Radius}
The right side of \Cref{fig:ablation} illustrates how increasing search radii have a diminishing effect on reconstruction quality. We further observe that at the smaller search radius of $0.03$ and a higher speed of $3.6^\circ$/frame, reconstruction quality begins to degrade over time. This affect is due to a divergence that occurs when the number neighbours around a particle drops below the necessary level to accurately represent the underlying geometry. We currently do not have a growing strategy that would detect such an event and multiply the particles in that region. This edge case and future work for implementing a growing strategy is further discussed in the supplementary. 

\vspace{-0.4cm}
\paragraph{Influence of Number of Particles}
In \Cref{fig:ablation} (left), we observed that increasing the number of particles used to initialize the scene asymptotically improves reconstruction quality, with $300,000$ particles producing the same quality as $200,000$. However, we also found that only approximately 7,000 particles contribute to regions with high density, with the remaining particles providing minimal contribution. The additional particles improve reconstruction quality in two ways: Firstly, they enable more particles to begin near the geometry, thereby reducing the initial convergence time. Secondly, as movement occurs, particles may be ejected from the geometry, and having free particles in empty regions enables the geometry to reabsorb these particles, thus maintaining reconstruction quality. These observations suggest the need for future work on developing a pruning and growing strategy for the particles.

\section{Conclusions, Limitations and Future Work}
\label{sec:LimAndConc}
\noindent\textbf{Limitations} Our particle embedding and online formulation require at least 10 views per frame to constrain the training, while offline dynamic methods that use time-conditioned deformation networks can use frames from any point during the trajectory to meet this requirement. We can overcome this disadvantage by incorporating depth supervision~\cite{abou2022implicit} and additional regularizers into the NeRF loss~\cite{kim2022infonerf}. 

ParticleNeRF currently has a cubic memory requirement -- albeit with a low constant -- which can be addressed by implementing a pruning strategy that removes particles with a density below a certain threshold. 

ParticleNeRF has lower visual fidelity on static scenes when compared to other methods. While we prioritized creating an encoding that is invariant to rigid transformations and quick to adapt to motions, we sacrificed the visual fidelity that comes with either multilevel encodings or positional Fourier transforms. These limitations are topics for future research.

\noindent\textbf{Conclusion} Our new particle-based parametric encoding, combined with the novel ability to backpropagate the rendering loss into the particle positions, enables fast adaptation to dynamic scenes. By leveraging the position of our particles, we have shown an alternative approach to using time-conditioned deformation networks for dynamic scenes. Our approach is quicker to adapt to movement in the scenes providing a consistent representation that is robust to movement.

{\small
\bibliographystyle{ieee_fullname}
\bibliography{egbib}
}

\clearpage

%%%%%%%%%% Merge with supplemental materials %%%%%%%%%%
%%%%%%%%%% Prefix a "S" to all equations, figures, tables and reset the counter %%%%%%%%%%
\setcounter{equation}{0}
\setcounter{figure}{0}
\setcounter{table}{0}
\setcounter{page}{1}
\setcounter{section}{0}
\renewcommand{\thesection}{S-\Roman{section}}
\renewcommand{\thetable}{S\arabic{table}}
\renewcommand{\theequation}{S\arabic{equation}}
\renewcommand{\thefigure}{S\arabic{figure}}
\makeatletter

%%%%%%%%%% Prefix a "S" to all equations, figures, tables and reset the counter %%%%%%%%%%

\onecolumn
\begin{center}
\textbf{\large Supplemental Materials: ParticleNeRF}
\end{center}

\section{Animated Blender Dataset}
We show a more detailed view of our results on the animated Blender dataset in \Cref{tab:synth}. Visuals can found in \Cref{fig:synth_vis}.

\section{Ablation Studies}
\begin{figure*}[t]
    \centering
    \includegraphics[width=0.8\linewidth]{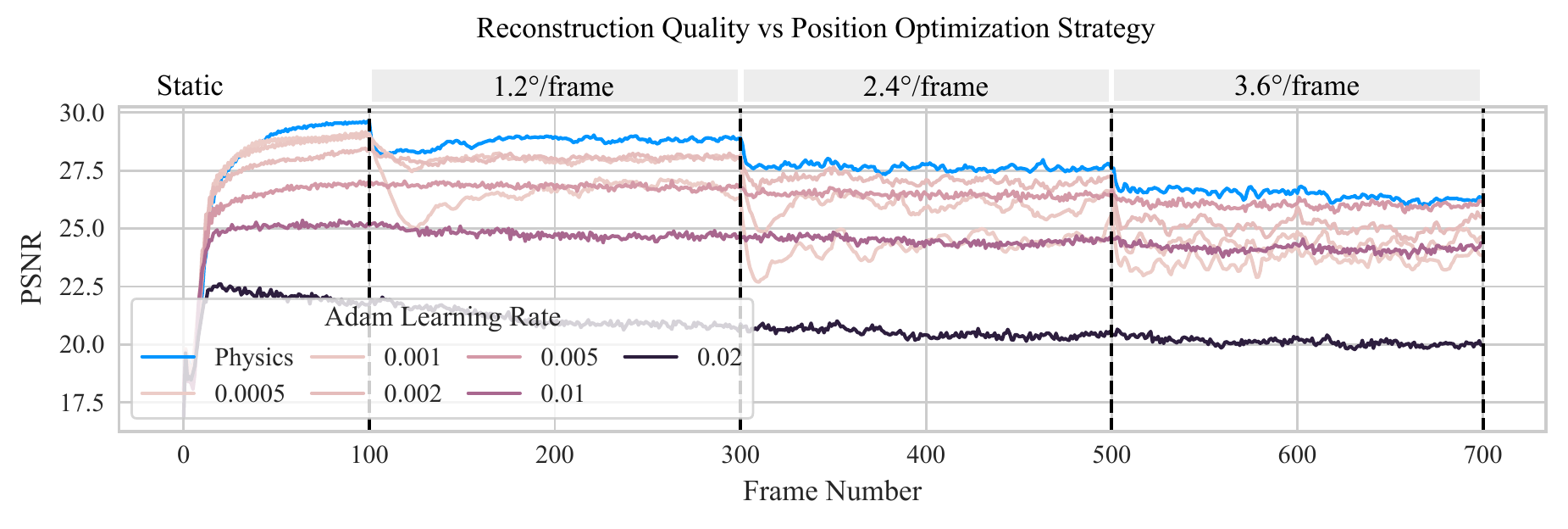}
    \caption{
    A comparison between using Adam and the physics system to update the particle positions after calculating their gradients relative to the NeRF reconstruction loss. Integrating the gradients with the physics system produces higher quality reconstructions on the wheel dataset whilst providing a well formulated means of adding constraints. Adam is configured with $\beta_1 = 0.9$, $\beta_2 = 0.99$ and a learning rate which is indicated in the legend. 
    }
    \label{fig:physics_ablation}
\end{figure*}

\textbf{Physics vs Adam} In \Cref{fig:physics_ablation}, we update the particle positions on the wheel dataset using Adam instead of using the PBD physics system. Our results show that the physics system produces higher quality reconstructions in dynamic scenes when compared to Adam. 

\begin{figure*}[t]
    \centering
    \includegraphics[width=0.8\linewidth]{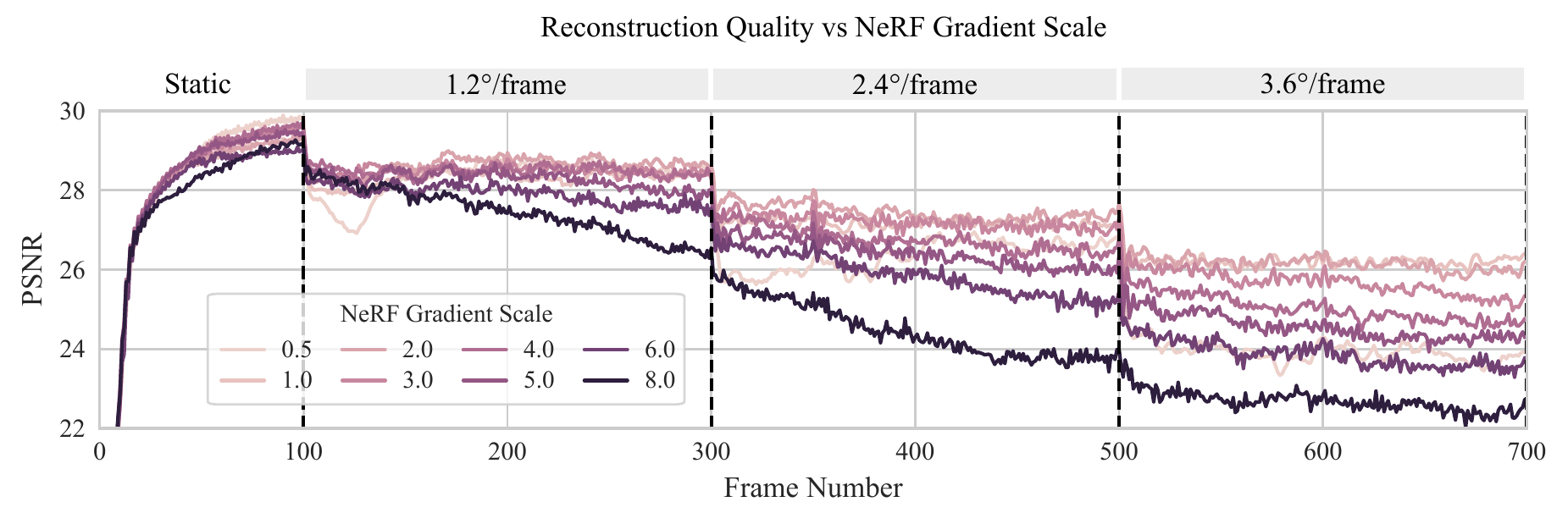}
    \caption{
    An ablation performed on the wheel dataset with 100,000 particles and a search radius of 0.04 showing the impact of the gradient scale $\alpha$ used to combine the NeRF gradients with the particle velocities. Values around 2.0 have the best results.
    }
    \label{fig:scale_ablation}
\end{figure*}

\textbf{Selecting the Gradient Scale} \Cref{fig:scale_ablation} shows how the gradient scale $\alpha$ -- used to integrate the positions gradients into the particle velocities -- was chosen for the wheel dataset. A small $\alpha$ creates a noticeable drop in quality when the wheel begins to rotate at frame 100 because the particle velocities take longer to adjust to the required speed. A large gradient scale creates particle instability. 

\begin{figure*}[]
    \centering
    \includegraphics[width=0.8\linewidth]{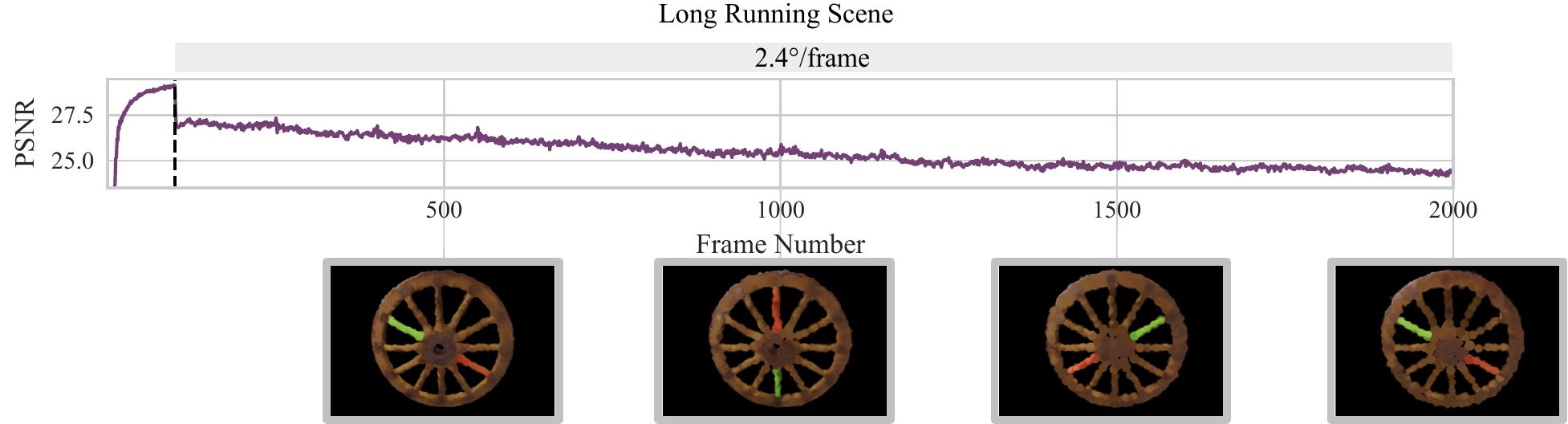}
    \caption{
    Figure showing degradation that occurs over a longer sequence of the Wheel dataset. The initial loss in quality is because the transition from static to 2.4 degrees is more sudden than the static to 1.2 degrees used in earlier experiments. Degradation occurs as particles begin to lose neighbours and are unable to acquire new ones.
    }
    \label{fig:long}
\end{figure*}

\textbf{Long Running Scene} In \Cref{fig:long}, the wheel experiment is run for around 2000 frames at $2.4^\circ\text{/second}$. We observe a degradation in quality over time as a result of particles slowly losing neighbours during the movement. When particles lose all their neighbours, a degenerate case occurs which inhibits other particles from reassembling within each other's search radius. This can be addressed in future work by creating a particle growing strategy that detects this case and does not allow lone particles to exist. Note, however, that even at the lowest reconstruction quality, the wheel maintains its shape. 

\begin{figure*}[htp]
    \centering
    \includegraphics[width=0.9\linewidth]{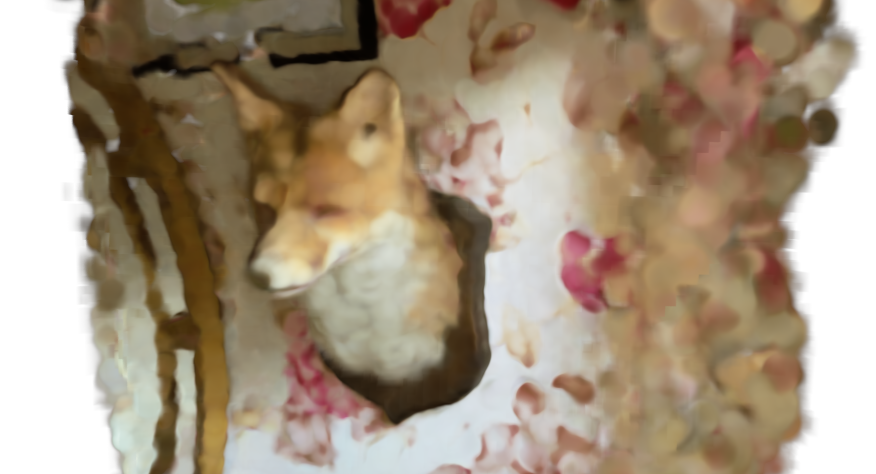}
    \caption{
    ParticleNeRF trained on a full scene after 3000 training steps with 200,000 particles at a search radius of 0.02. 
    }
    \label{fig:fox}
\end{figure*}

\textbf{Full Scene} We also visually demonstrate in \Cref{fig:fox} that ParticleNeRF can be used to reconstruct a full scene -- albeit at a reduced reconstruction quality when compared to InstantNGP.

\section{Implementation Details}
InstantNGP uses an occupancy grid as an acceleration structure to skip evaluating points in free space. In the original implementation, this structure was updated every 16 training steps by evaluating the density at each vertex in the grid. This worked well for static scenes because it was expected that the densities of the points would converge to their values and not change thereafter. In our dynamic scenes, we update the occupancy grid at every training step for both InstantNGP and ParticleNeRF. This allows the acceleration structure to quickly change with the scene and therefore does not inhibit training in areas which were once unoccupied but have since become occupied.

\textbf{Towards Realtime} In our dynamic dataset, out of the 100,000 particles initialized in the scene, only around 7000 are actually contributing to the geometry of the object. If we prune away particles with a density less than a certain threshold, our system can run a single training step in under \unit[6]{ms}. A pruning strategy however must be paired with a growing strategy so that particles can appear again when they are needed. A simple implementation of a growing strategy would be to replicate particles with less than a certain number of neighbours. The implementation of a robust pruning and growing strategy is left for future work.

\begin{figure*}[htp]
\centering
\includegraphics[width=0.95\linewidth]{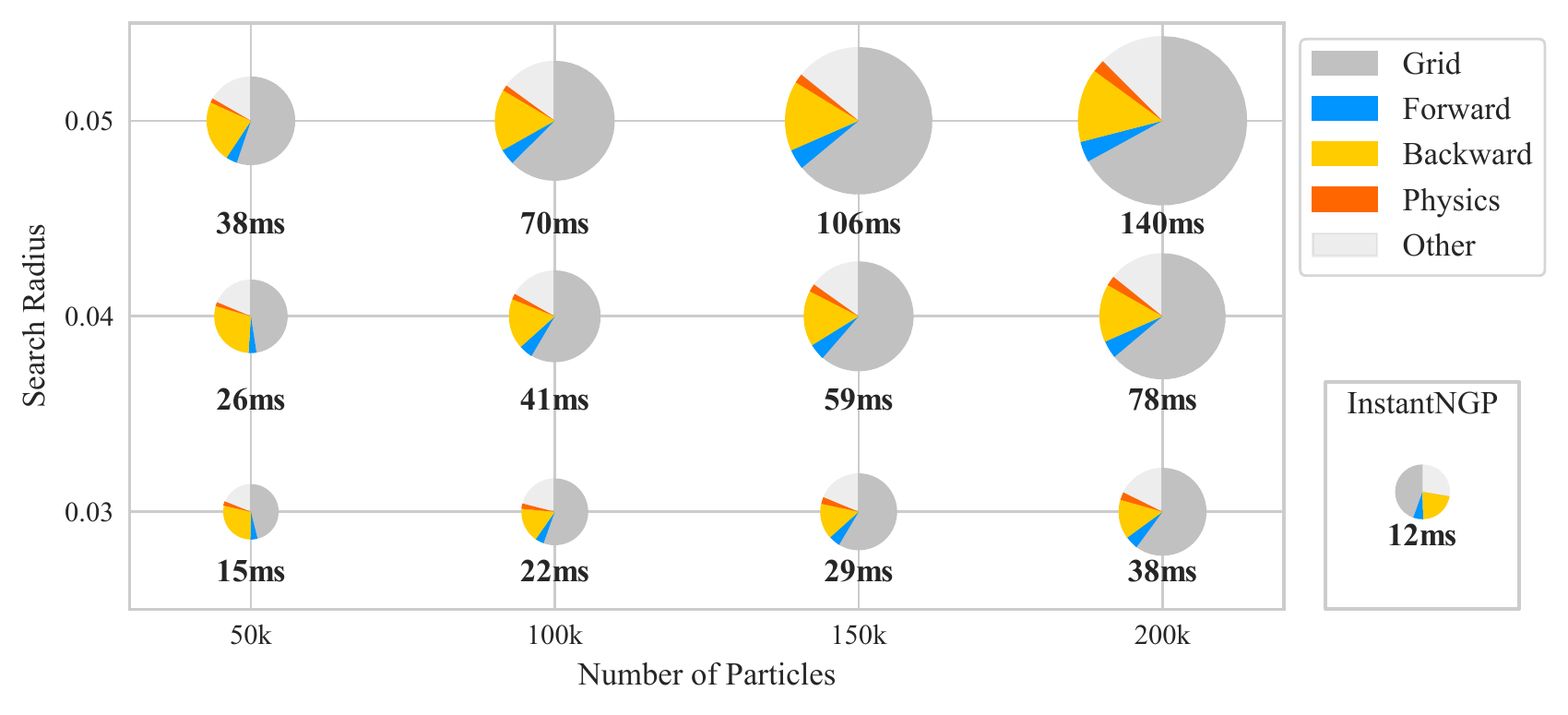}
\caption{
We profile ParticleNeRF and InstantNGP and show where the majority of the training time is spent per training step. The time taken by the acceleration structure (Grid), the forward pass, the backward pass, and the physics system are recorded for a range of configurations. InstantNGP is also shown on the right.
}
\label{fig:timings}
\end{figure*}

The computational cost of updating the acceleration structure at each training step is high, particularly for a ParticleNeRF with a large number of particles, as shown in \Cref{fig:timings}. To address this issue, an alternative acceleration structure that exploits the point-based nature of the encoding can be utilized. Z-buffering particles is another acceleration structure that can be used to skip free space. When a pruning strategy is implemented, it can be assumed that only the particles that produce geometry are present in the system. Therefore, Z-buffering can skip free space without requiring additional evaluations of the MLP, which is currently necessary.

Finally, we observe that the backward pass takes 5 times longer than the forward pass despite having a similar number of computations. The difference is due to the memory contention caused by multiple query points trying to update the same set of neighbouring particles. This too can be alleviated in future work. 

\begin{figure*}[t]
    \centering
    \includegraphics[width=\linewidth]{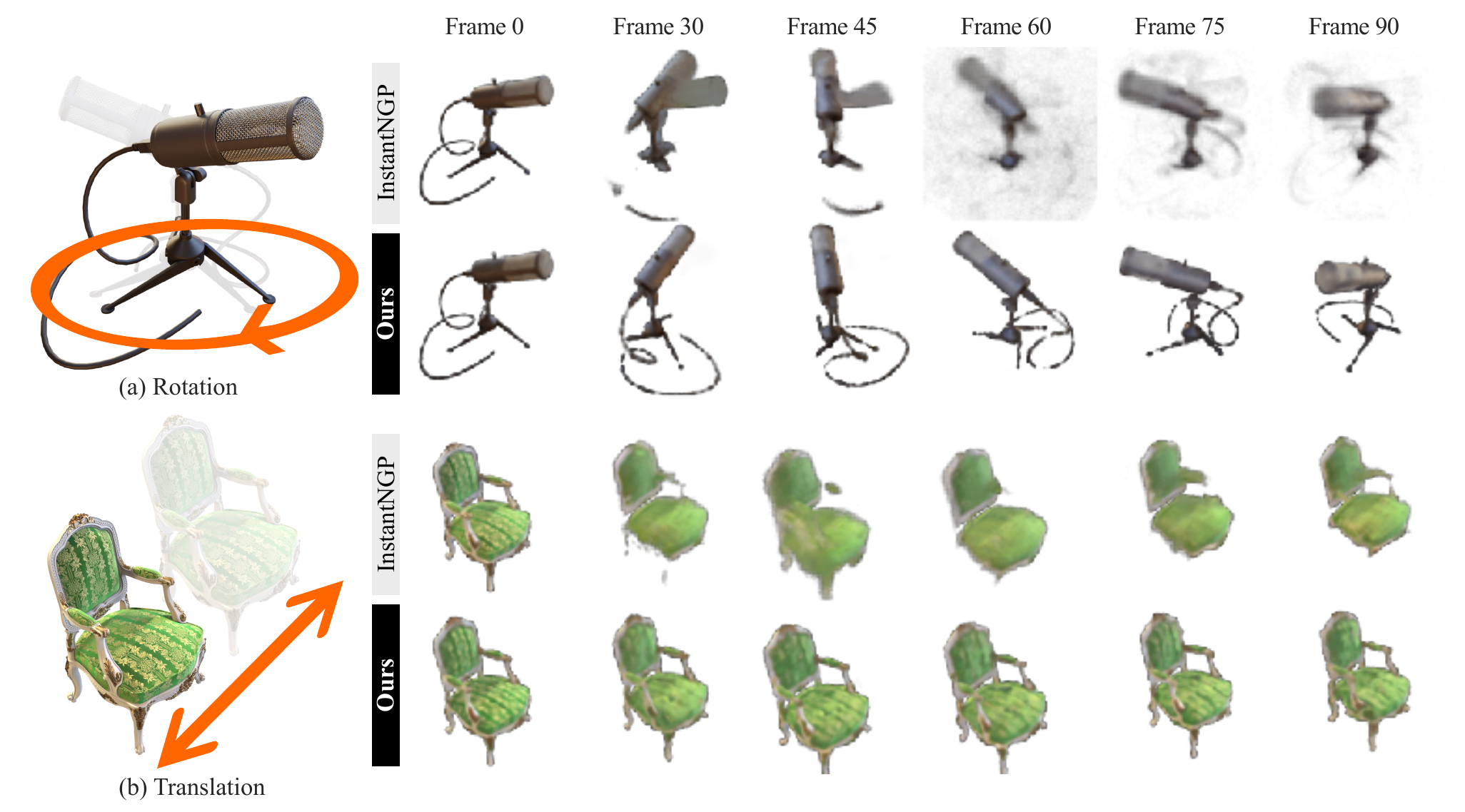}
    \caption{
    We test our encoding on an animated version of the Blender dataset. \textbf{(a)} shows an object rotating around its up axis. \textbf{(b)} shows an object translating from side to side. InstantNGP cannot learn features fast enough to maintain a high quality reconstruction. ParticleNeRF is able to move its features in space and maintain the structure of the object.
    }
    \label{fig:synth_vis}
\end{figure*}

\begin{table*}[b]
\centering
\caption{Performance of InstantNGP and ParticleNerf on the Animated Blender Dataset reported through the mean and standard deviation of the photometric PSNR over 100 frames.}

\resizebox{\linewidth}{!}{%
\label{tab:synth}
\rowcolors{2}{gray!25}{white} \cellcolor{white}
\tabcolsep=0.12cm
\begin{tabular}{lllllllllll}
\toprule
 & & Model & Chair & Drums & Ficus & Hotdog & Lego & Materials & Mic & Ship \\
     & \multirow[c]{-2}{*}{\makecell{Step\\per Frame}} & Encoding &  &  &  &  &  &  &  \\
\midrule
\cellcolor{white} & \cellcolor{white}& Hash & \textbf{27.43} & \textbf{21.82} & \textbf{26.40} & \textbf{30.13} & \textbf{27.00} & \textbf{25.25} & \textbf{29.61} & \textbf{22.24} \\
\multirow[c]{-2}{*}{\rotatebox[origin=c]{90}{Static}} & \multirow[c]{-2}{*}{} & Particle & 24.60 & 19.93& 22.92 & 27.35 & 23.23 & 23.33 & 28.06 & 23.09	 \\
\midrule
\cellcolor{white} & \cellcolor{white} & Hash & 21.30 $\pm$ 2.3 &	18.72 $\pm$ 1.4 &	22.44 $\pm$ 0.9 &	25.71 $\pm$ 1.1 &	21.07 $\pm$ 1.4 &	21.28 $\pm$ 1.5 &	26.42 $\pm$ 1.0 &	22.71 $\pm$ 1.1 \\
\cellcolor{white} & \multirow[c]{-2}{*}{1$^\circ$} & Particle & \textbf{24.02 $\pm$ 1.4} &	\textbf{19.70 $\pm$ 1.2} &	\textbf{22.85 $\pm$ 0.7} &	\textbf{27.02 $\pm$ 0.8} &	\textbf{22.77 $\pm$ 1.1} &	\textbf{23.23 $\pm$ 1.2} &	\textbf{27.77 $\pm$ 0.5} &	\textbf{23.04 $\pm$ 1.0} \\
\cellcolor{white} & \cellcolor{white} & Hash & 18.20 $\pm$ 1.9 &	17.76 $\pm$ 1.3 &	21.80 $\pm$ 0.9 &	24.13 $\pm$ 1.2 &	19.25 $\pm$ 1.4 &	18.80 $\pm$ 1.5 &	24.07 $\pm$ 1.0 &	22.15 $\pm$ 1.2 \\
\cellcolor{white} & \multirow[c]{-2}{*}{2$^\circ$} & Particle & \textbf{23.02 $\pm$ 1.4} &	\textbf{19.26 $\pm$ 1.2} &	\textbf{22.23 $\pm$ 0.7} &	\textbf{26.83 $\pm$ 0.8} &	\textbf{22.01 $\pm$ 1.1 }&	\textbf{22.65 $\pm$ 1.2} &	\textbf{26.97 $\pm$ 0.5} &	\textbf{22.93 $\pm$ 1.0} \\
\cellcolor{white} & \cellcolor{white} & Hash & 16.91 $\pm$ 1.7 &	16.92 $\pm$ 1.2 &	21.19 $\pm$ 0.9 &	23.22 $\pm$ 1.3 &	18.29 $\pm$ 1.5 &	17.20 $\pm$ 1.5 &	22.76 $\pm$ 1.4 &	21.64 $\pm$ 1.4  \\
\cellcolor{white} & \multirow[c]{-2}{*}{3$^\circ$} & Particle & \textbf{21.94 $\pm$ 1.5} &	\textbf{18.72 $\pm$ 1.3} &	\textbf{21.77 $\pm$ 0.8} &	\textbf{26.15 $\pm$ 0.8 }&	\textbf{21.32 $\pm$ 1.1} &	\textbf{22.11 $\pm$ 1.2} &	\textbf{26.32 $\pm$ 0.6} &	\textbf{22.64 $\pm$ 1.0}  \\
\cellcolor{white} & \cellcolor{white} & Hash & 16.56 $\pm$ 1.8 &	16.38 $\pm$ 1.0 &	20.66 $\pm$ 1.0 &	22.77 $\pm$ 1.4 &	17.65 $\pm$ 1.7 &	16.72 $\pm$ 1.5 &	22.99 $\pm$ 1.1 &	21.53 $\pm$ 1.3 \\
\cellcolor{white} \multirow[c]{-8}{*}{\rotatebox[origin=c]{90}{Rotation}} & \multirow[c]{-2}{*}{4$^\circ$} & Particle & \textbf{21.16 $\pm$ 1.5} &	\textbf{18.26 $\pm$ 1.2} &	\textbf{21.13 $\pm$ 0.8} &	\textbf{25.45 $\pm$ 0.9} &	\textbf{20.73 $\pm$ 1.2} &	\textbf{21.38 $\pm$ 1.3} &	\textbf{25.28 $\pm$ 0.7} &	\textbf{22.21 $\pm$ 1.1} \\
\midrule
\cellcolor{white} & \cellcolor{white}& Hash & 22.60 $\pm$ 2.1 &	19.06 $\pm$ 1.5 &	22.13 $\pm$ 0.8 &	25.18 $\pm$ 1.7 &	21.40 $\pm$ 1.6 &	22.87 $\pm$ 1.6 &	27.06 $\pm$ 1.1 &	21.89 $\pm$ 1.4  \\
\cellcolor{white} &  \multirow[c]{-2}{*}{$1\,\text{cm}$}  & Particle & \textbf{24.20 $\pm$ 1.4} &	\textbf{19.68 $\pm$ 1.3} &	\textbf{22.43 $\pm$ 0.7} &	\textbf{26.11 $\pm$ 1.5} &	\textbf{22.85 $\pm$ 1.1} &	\textbf{23.34 $\pm$ 1.3} &	\textbf{28.03 $\pm$ 0.6} &	\textbf{22.69 $\pm$ 1.2}  \\
\cellcolor{white} & \cellcolor{white} & Hash & 19.24 $\pm$ 2.1 &	17.68 $\pm$ 1.4 &	21.38 $\pm$ 0.9 &	22.97 $\pm$ 1.7 &	20.06 $\pm$ 1.6 &	21.07 $\pm$ 1.9 &	24.43 $\pm$ 1.3 &	21.41 $\pm$ 1.3 \\
\cellcolor{white} &  \multirow[c]{-2}{*}{$2\,\text{cm}$} & Particle & \textbf{23.15 $\pm$ 1.3} &	\textbf{19.31 $\pm$ 1.3} &	\textbf{21.60 $\pm$ 0.7} &	\textbf{25.38 $\pm$ 1.2} &	\textbf{22.07 $\pm$ 1.1} &	\textbf{23.06 $\pm$ 1.2} &	\textbf{27.12 $\pm$ 0.7} &	\textbf{22.71 $\pm$ 1.1}  \\
\cellcolor{white} & \cellcolor{white}  & Hash & 17.90 $\pm$ 2.0 &	17.12 $\pm$ 1.3 &	\textbf{21.11 $\pm$ 0.9} &	21.45 $\pm$ 1.8 &	19.37 $\pm$ 1.8 &	19.18 $\pm$ 2.3 &	23.47 $\pm$ 1.4 &	21.07 $\pm$ 1.4 \\
\cellcolor{white} \multirow[c]{-6}{*}{\rotatebox[origin=c]{90}{Translation}} &  \multirow[c]{-2}{*}{$3\,\text{cm}$} & Particle & \textbf{22.15 $\pm$ 1.4} &	\textbf{18.80 $\pm$ 1.2} &	20.94 $\pm$ 0.7 &	\textbf{24.28 $\pm$ 1.3} &	\textbf{21.29 $\pm$ 1.2} &	\textbf{22.69 $\pm$ 1.2} &	\textbf{26.16 $\pm$ 0.7} &	\textbf{22.15 $\pm$ 1.2}\\
\bottomrule
\end{tabular}
}
\end{table*}

 \section{$\text{SE}(3)$ Invariance and Interpretability}
 Our encoding is invariant under 3D rigid transformations. For a transformation $T \in \text{SE}(3)$:

  \begin{equation}
   \vf_j = F(\vx_j, \pcloud) = F(T\vx_j, T\circ\pcloud)
   \label{eq:invariance}
 \end{equation}
 where
  \begin{equation}
   T\circ\pcloud = \{(T\vx_i, T\vv_i, \vf_i) : i=1,2,..., M \}
   \label{eq:invariance2}
 \end{equation}

Our particle encoding approach generates local features that are directly associated with local geometry patches. As a result of its $\text{SE}(3)$ invariance, when these particles are moved or deformed, the underlying geometry will be similarly affected. For instance, if a subset of particles representing an object within a scene is moved or deformed, the corresponding geometry will change in predictable ways. In contrast, non-parametric NeRF methods encode geometry within the weights of a neural network. There is no natural transformation that can be applied to the weights that would change the scene in predictable ways. Additionally, voxel-based features, particularly those using a multilevel approach, cannot be easily transformed to modify the scene. The lack of controllability and interpretability over these features creates two challenges: (i) It is challenging to apply motion priors in dynamic scenes - for example in the case where a certain object's velocity is known. This limitation was also identified by the authors of TiNeuVox. (ii) The usefulness of NeRFs as a representation of geometry is limited when the underlying features lack the invariance property. Therefore, NeRFs without the invariance property are primarily suitable for view synthesis and not as a new method of geometry representation in downstream tasks.

\begin{figure}[h]
    \centering
    \includegraphics[width=\linewidth]{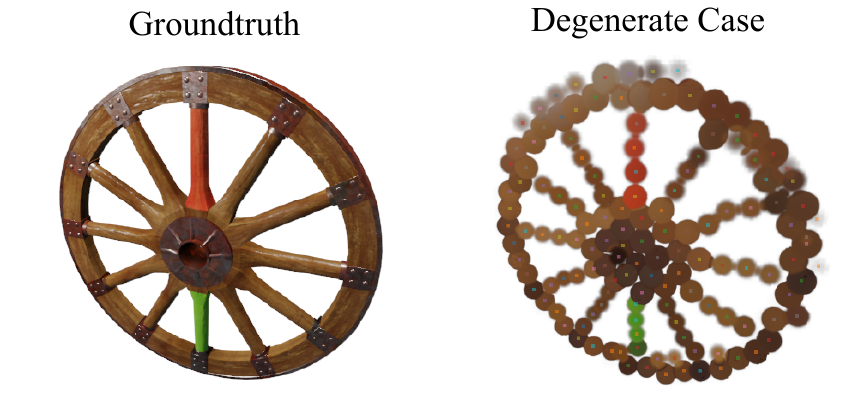}
    \caption{
    A degenerate case occurs when none of the particles have any neighbours within their search radius. The resulting reconstruction manifests as a set of spheres due to the inability of one particle alone to express more complicated geometries. 
    }
    \label{fig:degenerate}
\end{figure}

\section{Degenerate Case}
The interpolated features around a \textbf{lone} particle will be the same along the surface of a sphere centered on that particle. The effect is shown visually in \Cref{fig:degenerate} where we reduce the number of particles in the scene to 200 and ensure that every particle has no neighbours within its search radius. In our experiments, we used a sufficient number of particles when initializing the scene to circumvent the issue. Future work can develop a particle growing algorithm that detects lone particles and creates neighbours in their vicinity.

\end{document}